\documentclass[fullpaper]{nldl}

\usepackage[utf8]{inputenc}
\usepackage{url}
\usepackage{graphicx}
\usepackage{authblk}

\usepackage{algorithm}
\usepackage{algorithmic}
\usepackage{tikz}
\usepackage{comment}
\usepackage{amsmath,amssymb} 
\usepackage{color}

\usepackage[square,numbers]{natbib}

\usepackage{booktabs}
\usepackage{adjustbox}
\usepackage{multirow}
\usepackage{bbm}
\usepackage{makecell}
\usepackage{url}

\usepackage{breakurl}
\usepackage[breaklinks]{hyperref}
\usepackage{dblfloatfix}

\DeclareMathOperator*{\argmax}{argmax}
\DeclareMathOperator*{\softmax}{softmax}
\DeclareMathOperator*{\match}{match}
\DeclareMathOperator*{\pseudolabel}{pseudolabel}
\newcommand{\etal}{\textit{et al}.}

\usepackage{hyperref}

\title{Dense FixMatch: a simple semi-supervised learning method \\for pixel-wise prediction tasks}
\author[1,2]{Miquel Martí i Rabadán\thanks{Corresponding Author: miquelmr@kth.se}}
\author[2]{Alessandro Pieropan}
\author[1]{\\Hossein Azizpour}
\author[1]{Atsuto Maki}
\affil[1]{KTH Royal Institute of Technology, Stockholm, Sweden}
\affil[2]{Univrses AB, Stockholm, Sweden}

\date{\vspace{-3ex}}

\begin{document}
\nldlmaketitle

\begin{abstract} 
We propose Dense FixMatch, a simple method for online semi-supervised learning of dense and structured prediction tasks combining pseudo-labeling and consistency regularization via strong data augmentation.
We enable the application of FixMatch in semi-supervised learning problems beyond image classification by adding a matching operation on the pseudo-labels. This allows us to still use the full strength of data augmentation pipelines, including geometric transformations.

We evaluate it on semi-supervised semantic segmentation on Cityscapes and Pascal VOC with different percentages of labeled data and ablate design choices and hyper-parameters.
Dense FixMatch significantly improves results compared to supervised learning using only labeled data, approaching its performance with 1/4 of the labeled samples.

\end{abstract}

\section{Introduction}

Semi-supervised learning (SSL) has shown great potential to reduce the annotation costs of training deep learning models. Modern methods achieve competitive results at a fraction of the amount of annotated samples required for standard supervised learning~\cite{mixmatch,remixmatch,fixmatch}. The potential cost savings are even larger for structured or dense prediction tasks, such as object detection, instance or semantic segmentation since the annotation cost for such tasks is much larger than for image classification.

\begin{figure}[h]
    \centering
    \includegraphics[width=0.45\textwidth]{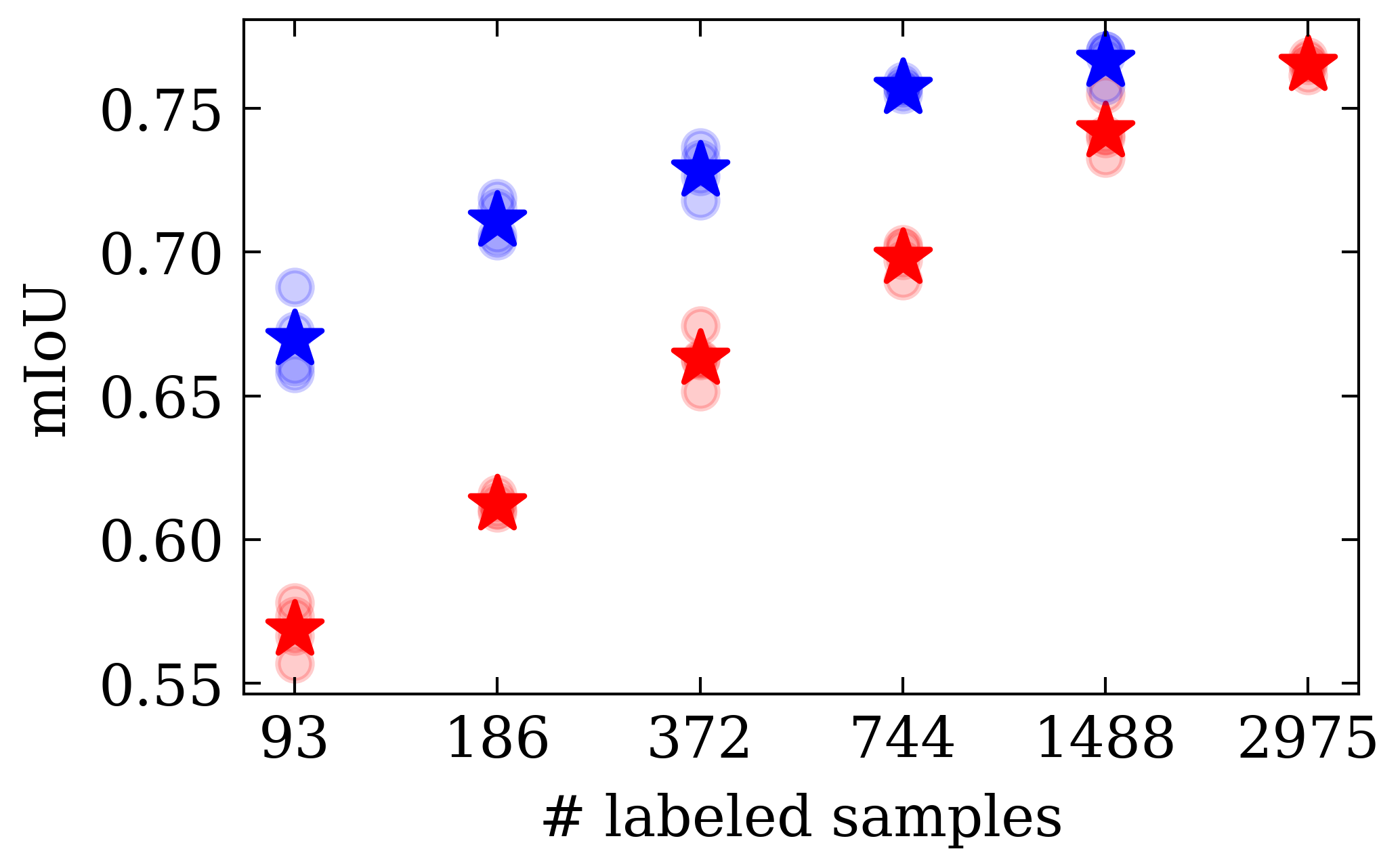}
    \caption{Dense FixMatch (blue) on unlabeled data improves the performance of semi-supervised semantic segmentation on Cityscapes \texttt{val} set using DeepLabv3+ with
    ResNet-101
    backbone over supervised baselines (red) across different amounts of labeled samples. $\star$ represents the mean over four different runs with random labeled data splits. Results for individual runs are shown with circles.}
    \label{fig:intro_fig}
    \vspace{-10pt}
\end{figure}

\begin{figure*}[h]
    \centering
    \includegraphics[width=\textwidth]{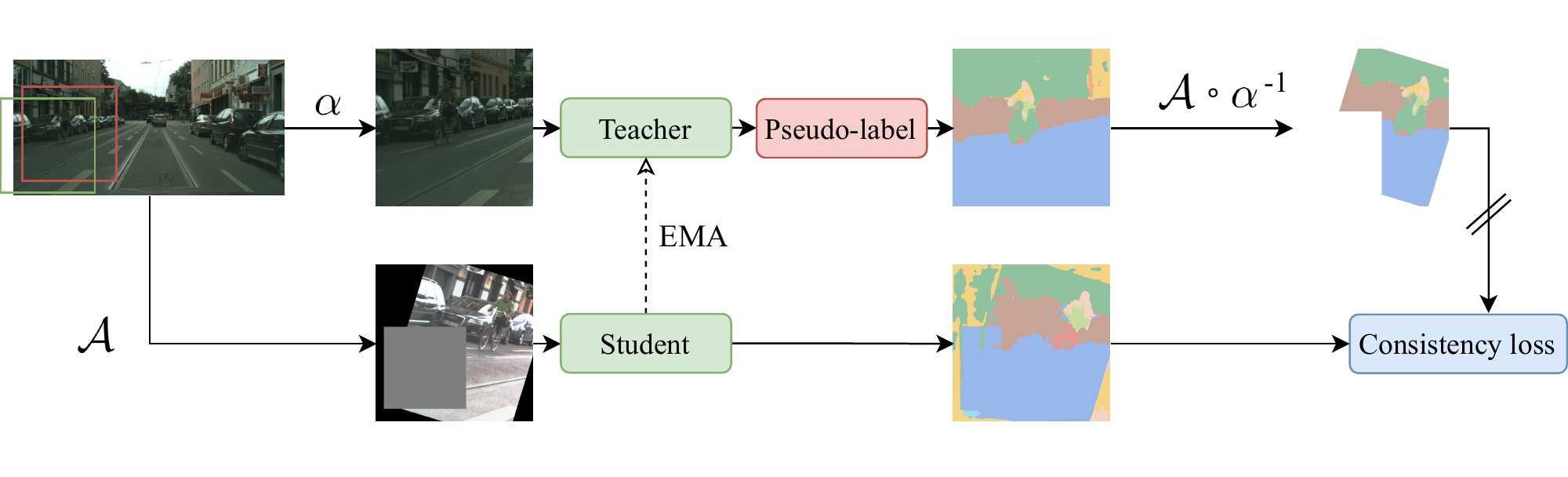}
    \caption{Dense FixMatch diagram for semantic segmentation. From an input image (top-left), two different views are created via $\alpha$ and $\mathcal{A}$, the weak and strong augmentation pipelines respectively. The squares represent the crops used for obtaining both views. Top: the first view is used by the teacher in the Mean Teacher~\cite{meanteacher} framework to generate pseudo-labels. These are matched to the second view after applying the inverse of the weak augmentation $\alpha^{-1}$, and then the strong one $\mathcal{A}$. Bottom: the second view is passed to the student model to obtain predictions to train against the pseudo-labels via the consistency loss, possible to define thanks to the shared structure and reference frame between both.}
    \label{fig:diagram}
\end{figure*}

However, SSL methods have been mainly developed and studied with image-level classification in mind~\cite{meanteacher,vat,noisy,mixmatch,fixmatch}. Only more recently, methods have appeared adapting or proposing solutions to structured or dense tasks such as object detection~\cite{rethinking,csd,stac,unbiasedteacher,softteacher} or semantic segmentation~\cite{cutmix_semi,cct,gct,pseudoseg,cps,ael}. Still, most works have focused on improving performance on specific tasks and not aimed at finding methods that could be applied to different tasks. Only a handful of methods are generic enough to be used for multiple tasks with no or few changes~\cite{cutmix_semi,noisy,rethinking,cps,stac}. Designing task-generic methods is important for ease of portability to new tasks and the goal of our work, as well as a must in multi-task learning scenarios.

To this end, we perform simple but effective modifications to FixMatch~\cite{fixmatch} to adapt it for a larger class of dense or structured task, staying as close as possible to the original formulation. We call our approach Dense FixMatch and summarise it in Figure~\ref{fig:diagram}. We align the reference frame of the pseudo-labels obtained from the weakly-augmented view with that of the predictions obtained from the strongly-augmented view.
This way, we can define a consistency loss at each output location while still using the full set of possible augmentations. Using strong and varied augmentations has been identified as a key component of self-training with input-consistency~\cite{analysis_selftraining,cutmix_semi} since they allow exploring larger neighbourhoods of the training data points in the input data manifold as well as in different directions. 
In addition, we incorporate the Mean Teacher (MT) framework so that learning is more robust to noisy pseudo-labels and imbalanced class size~\cite{unbiasedteacher,noisylabelstreasure}.

We evaluate our approach on semantic segmentation with Cityscapes and Pascal VOC 2012 datasets and show the results in Figure~\ref{fig:intro_fig} and Table~\ref{tab:few}, outperforming supervised baselines across different labeled data regimes by a large margin and achieving comparable results to other works in the literature.
We also compare different mini-batch sampling approaches to assess whether it is feasible to use our method for semi-supervised multi-task learning where separate labeled and unlabeled data sampling is not possible~\cite{analysis_fixmatch}.\\

Our contributions are as follows:
\begin{itemize}
    \item We propose Dense FixMatch, a simple method that adds a matching operation between pseudo-labels and predictions to FixMatch thereby enabling its use on semi-supervised learning for any dense or structured task.
    \item We study its performance on semi-supervised semantic segmentation on Cityscapes and Pascal VOC 2012, showing improvements across multiple labeled data regimes over supervised baselines. For Cityscapes, we get improvements of up to +0.1 mIoU for 93 and 186 labeled samples reaching \textbf{0.6697} mIoU and \textbf{0.7110} mIoU respectively, and +0.04 mIoU when using all labeled samples and extra unlabeled data reaching \textbf{0.8082} mIoU.
    \item We ablate our design choices and hyper-parameters to give practitioners insights on how to tune it for new tasks and datasets.
\end{itemize}

\section{Related work}
The success of semi-supervised learning~\cite{dssl_survey} has come mainly from its application to image classification with deep learning. Wei~\etal~\cite{analysis_selftraining} proved that SSL methods based on (a) self-training and (b) consistency regularization will achieve high accuracy with respect to ground-truth labels with the key to their success being to explore large enough neighbourhoods of the pseudo-labeled examples in the input data manifold, for example via aggressive data augmentation. Self-training or pseudo-labeling methods rely on bootstrapping current model predictions on unlabeled data and using them as labels. Consistency regularization relies on the assumption that small perturbations of the data points in either input or latent space should not change the output.

FixMatch~\cite{fixmatch} and Noisy Student self-training~\cite{noisy} are two methods combining such building blocks: the former follows an online approach where pseudo-labels are generated during training, and the latter has subsequent pseudo-labeling and re-training phases. Both use strong data augmentation to train against the pseudo-labels. Other works have also used other kinds of perturbations for the consistency objective, such as adversarial examples~\cite{vat}, network perturbations~\cite{pi,meanteacher,noisy} or MixUp~\cite{mixup,mixmatch} as well as other techniques to tackle distribution misalignment between true- and pseudo-labels~\cite{remixmatch}.

SSL applied to tasks other than image classification has also seen significant developments in recent years. For semi-supervised object detection, multiple works have used consistency regularization and perturbations via data augmentation~\cite{csd,stac,unbiasedteacher,softteacher}.
For semi-supervised semantic segmentation, the work in \cite{cutmix_semi} found that strong and varied perturbations are required and proposed CutMix~\cite{cutmix} as the strong augmentation. CCT~\cite{cct} enforces consistency between predictions perturbing latent features. GCT~\cite{gct} uses two differently initialized networks for co-training and a flaw detection module. CPS~\cite{cps} instead enforces consistency against hard pseudo-labels. Pseudoseg~\cite{pseudoseg} uses strong augmentation and fuses pseudo-labels from decoder predictions with ones from GradCAM~\cite{gradcam}. ST++~\cite{stpp} does self-training with strong data augmentation in the re-training phase while selecting and prioritizing reliable images. AEL~\cite{ael} focuses on balancing the performance between classes via different task-specific strategies. U2PL~\cite{u2pl} uses unreliable pseudo-labels for negative learning.

In contrast, we follow FixMatch as close as possible to keep the benefits of using online pseudo-labeling and consistency regularization between predictions on weakly and strongly augmented images. We add only a spatial matching operation to enable its use in dense and structured tasks and the MT framework for improving pseudo-label quality.

\section{Dense FixMatch}
We adapt FixMatch~\cite{fixmatch} for its use in structured and dense prediction tasks in the semi-supervised setting.

Our method assumes the standard framework of semi-supervised learning where labeled samples $X_L$ contribute to the supervised objective $\mathcal{L}_s$ and unlabeled samples $X_U$ are used in an unsupervised objective $\mathcal{L}_{u}$, with the option to use the labeled samples also for the latter. The unsupervised loss weight $\lambda$ trades off the contribution of both objectives to the final loss.

\begin{equation}
    \mathcal{L} = \mathcal{L}_s(\mathbf{x}, \mathbf{y}, \theta) + \lambda \mathcal{L}_{u}(\mathbf{x}, \theta)
\end{equation}

To define the unsupervised or consistency objective, FixMatch uses image-level pseudo-labels obtained from a weakly-augmented version of the unlabeled images (via augmentation pipeline $\alpha$) to supervise learning on the strongly-augmented version of the same images (via $\mathcal{A}$). For image classification, the output is expected to be invariant to the applied transformations and so the obtained pseudo-label can be directly used for this purpose.
In contrast, this is not possible when the output of the task at hand has a spatial structure related to that of the input and thus will vary depending on the applied augmentations. This is the case for dense or structured tasks such as semantic segmentation or object detection, among others. For those tasks, any geometric transformation of the input equivariantly transforms its corresponding output. Therefore, when using geometric transformations as part of the weak and strong augmentation pipelines, the obtained pseudo-labels will not generally match pixel-to-pixel or at each location.

We adopt a simple approach to align the predictions of one view (e.g. weak augmentation $\alpha$) to the reference frame of the other view (e.g. strong augmentation $\mathcal{A}$). Specifically, we first apply the inverse geometric transformation of the first view to the predictions obtained on it and then apply the geometric transformation of the second view so that predictions on both views end up in the same reference frame. This simple mechanism enables to define a consistency objective between the two views for any dense or structured task, including semantic segmentation, object detection, and instance segmentation, while still being able to use different geometric transformations in both augmentation pipelines. Figure~\ref{fig:diagram} illustrates our approach for the case of semantic segmentation.

We define the supervised objective $\mathcal{L}_s$ by applying a per-sample, location-wise objective $l_s$ between a prediction $\mathbf{\hat{y}}_i = f(\alpha(\mathbf{x}_i);\theta)$ by the model $f$ with weights $\theta$ on a weakly-augmented view of the labeled sample $\alpha(\mathbf{x}_i)$ and its corresponding (augmented) ground-truth label $\alpha(\mathbf{y}_i)$, and averaging over all valid locations $j$:

\begin{equation}
    \mathcal{L}_s = \frac{1}{B_L}\sum_{\mathbf{x}_i\in\mathcal{B}^l}\frac{1}{S_i}\sum_{j=1}^{S_i}l_s(\mathbf{\hat{y}}_{i,j}, \alpha(\mathbf{y}_i)_{j}),
\end{equation}

\noindent where $\mathcal{B}^l$ is the part of the mini-batch corresponding to labeled samples with size $B_L$, and $S_i$ is the number of valid locations in the augmented ground-truth label $\alpha(\mathbf{y}_i)$, e.g. for semantic segmentation the total number of valid pixels, width $\times$ height. The label corresponding to location $j$ is denoted by $\mathbf{y}_{i,j}$.

The unsupervised or consistency objective $\mathcal{L}_{u}$ is then defined by applying the same or a different location-wise objective $l_{u}$ between the predictions of a strongly-augmented view $\mathbf{\tilde{y}}_{i}=f(\mathcal{A}(\mathbf{x}_i);\theta)$ and the pseudo-label $\mathbf{\bar{y}}_{i}$
obtained from a weakly-augmented view of the same sample after spatially matching them:

\begin{equation}
    \mathcal{L}_u = \frac{1}{B_U}\sum_{\mathbf{x}_i\in\mathcal{B}^u}\frac{1}{S_i}\sum_{j=1}^{S_i}l_u(\mathbf{\tilde{y}}_{i,j},\match(\mathbf{\bar{y}}_{i}; \alpha, \mathcal{A})_j),
\end{equation}

\noindent where, $\mathbf{\bar{y}}_i=\pseudolabel(\mathbf{\hat{y}}_i)$ is a pseudo-label created from prediction $\mathbf{\hat{y}}_{i}$,
$\match()$ is the matching operation applied to spatially align pseudo-label $\mathbf{\bar{y}}_i$ with prediction $\mathbf{\tilde{y}}_i$,
$\mathcal{B}^u$ is the part of the mini-batch corresponding to unlabeled samples with size $B_U$,
and $S_i$ is the number of valid locations in the matched pseudo-labels.

For our experiments on semantic segmentation, we adopt the cross-entropy loss for both $l_s$ and $l_u$. We describe next the details of the $\match$ and $\pseudolabel$ operations, the latter being dependent on the task. Note that, for the unsupervised loss $\mathcal{L}_u$, gradients are back-propagated only through the predictions on strongly-augmented samples $\mathbf{\tilde{y}}$, and not through the pseudo-labels $\mathbf{\bar{y}}$.\\

\noindent\textbf{From prediction to pseudo-label.} Depending on the task at hand, obtaining pseudo-labels from the model predictions requires some post-processing that is usually done outside of the model. For our experiments in semantic segmentation, the model outputs normalized classification probabilities, using $\softmax$ operation, for each possible class at each output location. Obtaining a ``hard'' pseudo-label means retaining only the most likely class at each location, which is achieved by applying the $\argmax$ operation. In addition, it is common to use only high-confidence predictions as pseudo-labels~\cite{noisy,fixmatch}. To do so we define a confidence threshold $\tau$ above which to retain the labels. Locations with predictions of confidence below the threshold will not have valid corresponding pseudo-labels and thus will not contribute to the loss:

\begin{equation}
\begin{split}
    \pseudolabel(\mathbf{\hat{y}}_{i,j}; \tau) = \\
     \mathbbm{1}(\max(\mathbf{\hat{y}}_{i,j})\geq\tau)\argmax(\mathbf{\hat{y}}_{i,j};\theta)).
\end{split}
\end{equation}\\

\noindent\textbf{Matching operation.} There are multiple options to spatially align or match predictions and pseudo-labels. One could achieve so by (a) bringing both to the original image reference frame inverting the transforms applied to both views; (b) bringing the prediction to the pseudo-label reference frame by applying to the prediction the inverse of the geometric transform in the strong augmentation pipeline and the geometric transform in the weak augmentation pipeline; or (c) bringing the pseudo-label to the prediction reference frame by first applying the inverse of the geometric transforms in the weak augmentation pipeline, $\alpha^{-1}$, and then the transforms of the strong augmentation pipeline, $\mathcal{A}$, to the pseudo-label $\mathbf{\bar{y}}_i$.
We choose the latter since it importantly does not operate on the predictions, only on the pseudo-labels, and thus does not require to back-propagate gradients through the matching operation:
\begin{equation}
    \match(\mathbf{\bar{y}}_{i}; \alpha, \mathcal{A}) = \mathcal{A}(\alpha^{-1}(\mathbf{\bar{y}}_{i})).
\end{equation}
It is important to note that applying both $\alpha^{-1}$ and $\mathcal{A}$ might produce invalid results for some locations of the pseudo-labels, e.g. the inverse of a random crop will place the pseudo-labels on the original location of the original sample view but areas outside the crop will not be defined. Such areas with invalid pseudo-labels must be tracked so that predictions on them do not contribute to the loss.\\

\noindent\textbf{Augmentation pipelines.} Dense FixMatch, like FixMatch, uses two different augmentations pipelines for generating the pseudo-labels and for training against them. Pseudo-labels are generated via the weak augmentation pipeline $\alpha$, while predictions to train on the pseudo-labels are obtained via the strong augmentation pipeline $\mathcal{A}$. FixMatch~\cite{fixmatch} applied to image classification uses a standard flip-and-shift augmentation strategy as $\alpha$ and either RandAugment~\cite{randaug} or CTAugment~\cite{remixmatch} as $\mathcal{A}$, always applying Cutout~\cite{cutout} last. We adopt similar choices for semantic segmentation, choosing RandAugment~\cite{randaug} for the strong augmentation $\mathcal{A}$ for simplicity, including both geometric and color transforms. We use random crops for both pipelines before applying the rest of transforms to ensure the same input size and instead of the shift operation in $\alpha$ since both achieve a similar effect. RandAugment randomly samples two transformations from a list of augmentations and a random magnitude for each within a range. The pool of augmentations includes both geometric and color transforms and we study the importance of using both types in Section~\ref{sec:ablations}. Instead, other works focusing on dense tasks and inspired by FixMatch avoid the misalignment by dropping the geometric transforms~\cite{pseudoseg,unbiasedteacher}, applying the same to both views~\cite{pixmatch}, or applying them only as part of $\mathcal{A}$~\cite{softteacher}.\\

\noindent\textbf{Mean Teacher.} In order to obtain cleaner and more stable pseudo-labels~\cite{noisylabelstreasure,unbiasedteacher}, we use the teacher in  MT~\cite{meanteacher} instead of the same model to generate pseudo-labels on the weakly-augmented views: $\mathbf{\tilde{y}}_i=f(\alpha(\mathbf{x}_i);\bar{\theta})$. The teacher weights $\bar{\theta}$ are updated after each training step with an exponential moving average (EMA) of the student weights according to Eq.~\ref{eq:ema}, with decay rate $m$. We ablate this choice in Section~\ref{sec:ablations}.

\begin{equation}
    \bar{\theta} \xleftarrow{} m\bar{\theta}+(1-m)\theta
    \label{eq:ema}.
\end{equation}\\

\section{Experiments}
\textbf{Datasets.} We use Cityscapes~\cite{cityscapes} and Pascal VOC 2012~\cite{voc} datasets for evaluating our method on semi-supervised semantic segmentation. Cityscapes consists of 2975 samples for training, with fine annotations for 19 classes, and 500 samples for evaluation. In addition, further 20000 samples are available in the \texttt{extra} set with coarse annotations, but we will use them later as unlabeled samples only. Pascal VOC consists of 1464 samples for training in the original set with annotations for 21 classes including background, and 1449 samples for evaluation in the validation set. Moreover, there are further 9118 labeled samples in the augmented set from SBD~\cite{sbd}. As is common in the literature, we simulate the semi-supervised setting with labeled and unlabeled splits of the training set with different labeled data regimes or ratios of labeled samples. For each split, we generate four different random splits with no guarantees of class balance. For Pascal VOC, we split the original set and use the augmented set as unlabeled data.

\begin{table*}
\adjustbox{max width=\textwidth}{
    \centering
    \begin{tabular}{lcccccccc}
    \toprule
          Method & Backbone & Sampling & \textbf{93}$_{(1/32)}$ & \textbf{186}$_{(1/16)}$ & \textbf{372}$_{(1/8)}$ & \textbf{744}$_{(1/4)}$ & \textbf{1488}$_{(1/2)}$ & \textbf{2975}$_{All}$ \\
         \midrule
         \multirow{2}{*}{Supervised} & RN-50 & & $.5579_{\pm.0091}$ & $.6004_{\pm.0012}$ & $.6550_{\pm.0051}$ & $.6943_{\pm.0065}$ & $.7332_{\pm.0095}$ & $.7608_{\pm.0054}$ \\
         \cmidrule{2-9}
          & RN-101 & &  $.5686_{\pm.0080}$ &$.6122_{\pm.0025}$ & $.6628_{\pm.0081}$ & $.6979_{\pm.0049}$ & $.7421_{\pm.0081}$ & $.7652_{\pm.0023}$ \\

         \midrule
         \multirow{4}{*}{\makecell[l]{Dense \\ FixMatch}} & \multirow{2}{*}{RN-50} & Explicit & $.6581_{\pm.0202}$ &  $.7013_{\pm.0079}$&  $.7243_{\pm.0049}$&  $.7504_{\pm.0063}$&  $.7599_{\pm.0063}$&  \\
         && Implicit & $.6554_{\pm.0158}$ & $.7065_{\pm.0065}$& $.7339_{\pm.0055}$& $.7547_{\pm.0070}$&$.7637_{\pm.0079}$\\
         \cmidrule{2-8}
         & \multirow{2}{*}{RN-101 } & Explicit & $.6697_{\pm.0119}$ &  $.7110_{\pm.0061}$&  $.7283_{\pm.0069}$& $.7572_{\pm.0015}$&  $.7666_{\pm.0050}$ \\
         &  & Implicit & $.6481_{\pm.0178}$ &  $.7083_{\pm.0098}$&  $.7391_{\pm.0028}$& $.7565_{\pm.0058}$&  $.7635_{\pm.0088}$ \\

    \bottomrule
    \end{tabular}}
    \caption{Results of Dense FixMatch on \textbf{Cityscapes} \texttt{val} set with few labeled samples on different amounts of labeled data, ResNet-50/101 backbones and DeepLabv3+, and either explicit or implicit mini-batch sampling settings. Dense FixMatch significantly improves over the baselines for both settings.
    }
    \label{tab:few}
    \vspace{-5pt}
\end{table*}
\begin{table*}
\adjustbox{max width=\textwidth}{	
    \centering	
    \begin{tabular}{lcccccccc}	
    \toprule	
          Method & Backbone & Sampling & \textbf{92}$_{(1/16)}$ & \textbf{183}$_{(1/8)}$ & \textbf{366}$_{(1/4)}$ & \textbf{732}$_{(1/2)}$ & \textbf{1464}$_{Original}$ & \textbf{10582}$_{Augmented}$ \\	
         \midrule	
         \multirow{2}{*}{Supervised} & RN-50 & & $.4075_{\pm.0114}$ & $.5361_{\pm.0257}$ & $.6183_{\pm.0153}$ & $.6788_{\pm.0056}$ & $.7214_{\pm.0029}$ & $.7522_{\pm.0038}$ \\	
         \cmidrule{2-9}	
          & RN-101 & &  $.4482_{\pm.0256}$ &$.5771_{\pm.0247}$ & $.6534_{\pm.0081}$ & $.7059_{\pm.0041}$ & $.7454_{\pm.0023}$ & $.7722_{\pm.0017}$ \\	
         \midrule	
         \multirow{4}{*}{\makecell[l]{Dense \\ FixMatch}} & \multirow{2}{*}{RN-50} & Explicit & $.5215_{\pm.0246}$ &  $.6249_{\pm.0374}$&  $.6902_{\pm.0045}$&  $.7169_{\pm.0010}$&  $.7391_{\pm.0012}$&  \\	
         && Implicit & $.4928_{\pm.0284}$ &  $.5892_{\pm.0334}$&  $.6729_{\pm.0076}$&  $.7031_{\pm.0028}$&  $.7432_{\pm.0045}$&  \\	
         \cmidrule{2-8}	
         & \multirow{2}{*}{RN-101 } & Explicit & $.5485_{\pm.0501}$ &  $.6582_{\pm.0334}$&  $.7204_{\pm.0059}$& $.7473_{\pm.0008}$&  $.7716_{\pm.0020}$ \\	
         &  & Implicit & $.4984_{\pm.0300}$ &  $.6133_{\pm.0313}$&  $.7047_{\pm.0053}$& $.7414_{\pm.0044}$&  $.7710_{\pm.0046}$ \\	
    \bottomrule	
    \end{tabular}}	
    \caption{Results of Dense FixMatch on \textbf{Pascal VOC 2012} \texttt{val} set with few labeled samples on different amounts of labeled data, ResNet-50/101 backbones and DeepLabv3+, and explicit or implicit settings.
    }	
    \label{tab:few_voc}
    \vspace{-10pt}
\end{table*}

\noindent \textbf{Model.} Following the literature, we use DeepLabv3+~\cite{deeplabv3} on ResNet-50 or ResNet-101 backbones~\cite{resnet} to define our semantic segmentation model, giving predictions at the same spatial resolution as the input. The model is initialized with ImageNet~\cite{imagenet} pre-trained weights.

\noindent \textbf{Implementation details.} We implement and train our models using PyTorch~\cite{pytorch} with distributed training and mixed precision on up to four NVIDIA A100 GPUs, depending on the total batch size and the backbone used. We use the computer vision library Kornia~\cite{kornia} for implementing the data augmentation pipelines for its support of invertible geometric transforms and differentiable augmentations for multiple tasks, including semantic segmentation. We follow EMAN~\cite{eman} and use the exponential moving average of the Batch Normalization statistics of the student to update the teacher.

\noindent \textbf{Evaluation.} We evaluate the performance of our method using as the main metric the mean Intersection-over-Union (mIOU) over all classes. For simplicity, we use full-resolution, single-scale, and single-pass evaluation in contrast to the sliding evaluation approach used in other works~\cite{cps,ael,u2pl}. For stability, the model used for evaluation has weights following the EMA of the weights obtained during training. For most experiments, this means exactly using the teacher in the MT framework for evaluation. For each labeled data regime, we train our model on each of the four random data splits using also a different random seed for each training run. We take the best checkpoint of each run according to the mIOU and compute the mean and standard deviation over the four runs.

\noindent \textbf{Training details.} We follow the training details of \cite{u2pl}. We employ two different mini-batch sampling strategies for SSL: (a) the common \textit{explicit} setting in which labeled and unlabeled data are sampled separately, and (b) the alternative \textit{implicit} setting in which all data is sampled uniformly regardless of labels~\cite{analysis_fixmatch}.
We train each setting for the equivalent of 80 or 240 epochs on the full train set for the supervised baseline, i.e. 52910 or 89250 updates, for Pascal VOC and Cityscapes, respectively.

\subsection{Results on few labeled samples}
In Figure \ref{fig:intro_fig} and Tables \ref{tab:few} and \ref{tab:few_voc}, we show results for the common splits of 1/32 to 1/2 of all the full \texttt{train} set for Cityscapes and 1/16 to the full original \texttt{train} set for Pascal VOC 2012, respectively. We compare few-supervision baselines using only the labeled data with Dense FixMatch, which also uses the unlabeled samples in either the explicit or implicit settings. Our method performs better than the baselines for both mini-batch sampling approaches. The explicit setting is slightly better for more label-scarce regimes but both give similar results with more labeled data.

\begin{table*}[t]
    \centering
    \begin{minipage}{.67\textwidth}
        \adjustbox{max width=\columnwidth}{
    \centering
    \begin{tabular}{lccccc}
    \toprule
          Method & Backbone & & & \multicolumn{2}{c}{mIoU} \\
         \midrule
         \multirow{2}{*}{Supervised} & RN-50 & &  & \multicolumn{2}{c}{$.7608_{\pm.0054}$} \\
         \cmidrule{2-6}
          & RN-101 & &  & \multicolumn{2}{c}{$.7652_{\pm.0023}$}  \\

         \midrule
         & & \texttt{train}  & \texttt{extra} & Explicit & Implicit \\
         \cmidrule{2-6}
         \multirow{6}{*}{\makecell[l]{Dense \\ FixMatch}} & \multirow{3}{*}{RN-50} & \checkmark &  &  \multicolumn{2}{c}{$.7869_{\pm.0018}$}\\
         & & & \checkmark& $.7916_{\pm.0026}$ & $.7935_{\pm.0017}^\dagger$\\
         & & \checkmark & \checkmark & $\mathbf{.7998_{\pm.0020}}$ & $.7948_{\pm.0016}$ \\
         \cmidrule{2-6}
         & \multirow{3}{*}{RN-101} & \checkmark &  & \multicolumn{2}{c}{$.7907_{\pm.0020}$}\\
          && & \checkmark & $.8005_{\pm.0010}$ & $.7974_{\pm.0020}^\dagger$\\
          && \checkmark & \checkmark  & $\mathbf{.8082_{\pm.0024}}$ & $.8012_{\pm.0013}$\\
    \bottomrule
    \end{tabular}}
    \caption{\small{Results on Cityscapes full labeled set and \texttt{extra} unlabeled samples for the supervised baselines and semi-supervised Dense FixMatch in both the explicit and implicit mini-batch sampling settings for SSL. We also compare using our method as a regularization on the labeled data only and using it on both labeled and unlabeled data. $^\dagger$is our setting for the main experiments.}}
    \label{tab:extra}
    \vspace{-10pt}
    \end{minipage}
    \hspace{1pt}
    \begin{minipage}{.3\textwidth}
    \centering 
    \adjustbox{max height=100pt}{
    \begin{tabular}{ccc}
        \toprule
        $\tau$ & $\lambda$ & mIoU\\
        \midrule
        \multirow{5}{*}{0} & 0.2 & 0.6566 \\
        & 0.5 & 0.6464 \\
        & 1 & \textbf{0.6594}\\
        & 2 & \textit{0.6581} \\
        & 5 & 0.5699\\
        \midrule
        \multirow{7}{*}{0.5} & 0.1 & 0.6447\\
          & 0.2 & \textbf{0.6609}\\
         & 0.5 & 0.6495\\
         & 1 & 0.6501$^\dagger$\\
          & 2 & \textit{0.6554}\\
         & 5 & 0.6090\\
          & 10 & 0.5276\\
         \midrule
         0.8 & 1 & 0.6388\\
         \midrule
         0.95 & 1 & 0.6148\\
         \bottomrule \\
    \end{tabular}
    }
    \vspace{-10pt}
    \caption{\scriptsize{Ablation study on the pseudo-label confidence threshold $\tau$ and the consistency loss weight $\lambda$. $^\dagger$values used in the main experiments.}}
    \label{tab:conf_threshold_weight_ablation}
    \end{minipage}
\end{table*}

\begin{table}[h!]
        \centering
        \adjustbox{max width=\columnwidth}{
        \begin{tabular}{ccccc}
        \toprule
        Method &&&& mIoU \\
        \midrule
        Supervised & & & & 0.5409 \\
        \multicolumn{2}{l}{Mean Teacher$^{\ast}$}& & & 0.5820 \\
        \midrule
          & Crop relation & Augmentation& MT & \\
         \cmidrule{2-5}
         \multirow{11}{*}{\makecell[l]{Dense \\ FixMatch}} &  Same & Crop+color&$\checkmark$ & 0.5794   \\
          & Same & Crop+color+cutout&$\checkmark$ & 0.6531 \\
          & Min. overlap & Crop+color+cutout&$\checkmark$ &  0.6542 \\
          & Min. overlap & Crop+geom. &$\checkmark$& 0.6539 \\
          & Min. overlap & Crop+geom.+cutout &$\checkmark$& 0.6517 \\
          & Min. overlap & Crop+color+geom. &$\checkmark$& 0.6579 \\
          & Same & Crop+color+geom. &$\checkmark$& 0.6157 \\
          & Any & Crop+color+geom.+cut. &$\checkmark$& 0.6300  \\
         & Same & Crop+color+geom.+cut. &$\checkmark$& \textbf{0.6660} \\
          & Min. overlap & Crop+color+geom.+cut. &$\checkmark$& \textit{0.6594}$^\dagger$ \\
          & Min. overlap & Crop+color+geom.+cut. & & 0.6275 \\
         \bottomrule \\
    \end{tabular}}
    \caption{\small{Ablation on the use of Mean Teacher (MT) framework, the relation between crops in the two views of Dense FixMatch and the use of the geometric, color or all augmentations in RandAugment. $^{\ast}$Using MT on its own with the augmentation pipeline of the supervised baseline. We use a logistic warm-up schedule for the consistency weight during the first 60 epochs. $^\dagger$is our setting for the main experiments.}}
    \label{tab:further_ablation}
\end{table}

\subsection{Results on full labeled set and extra unlabeled samples}
In addition, we evaluate Dense FixMatch in the more realistic setting where all labeled samples are used and extra unlabeled data is available in Cityscapes. We use all samples from the \texttt{extra} set but discard the coarse annotations and just treat them as unlabeled samples. We also compare to the fully-supervised baseline using only the labeled data and using the consistency loss as a regularization term only, i.e. computed on the same labeled data as the supervised loss. Results are shown in Table \ref{tab:extra}. Computing the loss on both labeled and unlabeled samples gives the best results.

\subsection{Ablations}
\label{sec:ablations}
In Table~\ref{tab:conf_threshold_weight_ablation}, we ablate the main hyper-parameters of our method using the 93 labeled samples regime of semi-supervised semantic segmentation with Cityscapes, with a single data split and seed, and the explicit setting.
For dense tasks such as semantic segmentation, other works have reported that using a low $\tau$ or removing it altogether gives better final results~\cite{pseudoseg}, hypothesizing that discarding predictions of lower confidence makes the loss dominated by easy classes~\cite{pixmatch,ael}. Moreover, low-confidence predictions in semantic segmentation tend to concentrate in truly ambiguous regions such as the boundaries between objects of different classes~\cite{u2pl} so discarding them means removing supervision mainly from boundary pixels which are in fact the most informative. Values for $\lambda$ between 0.2 and 2 give comparable results.

In Table~\ref{tab:further_ablation} we ablate our design choices: the relation between crops in both views, the choice of augmentations, and the use of MT. The best results are obtained when using MT and all possible augmentations, although using the same crop instead of overlapping crops between views improves results. We hypothesize this is due to a larger number of valid locations in the matched pseudo-labels.

\begin{table*}[h!]
    \centering
    \adjustbox{max width=\textwidth}{
    \begin{tabular}{ccccccccccccccccccccc}
         \toprule
         \textbf{Method} & Road & Side. & Build. & Wall & Fence & Pole & T.light & T.sign & Veg. & Terr. & Sky & Person & Rider & Car & Truck & Bus & Train & Motor. & Bic. & \textbf{Mean}\\
         \midrule
         \% pixels & 36.02 &  7.06 &  25.61 &  0.41 &  0.81 &  1.16 &  0.18 &  0.59 &  14.08 &  1.22 &  3.80 &  1.08 &  0.09 &  6.73 &  0.25 &  0.23 &  0.21 &  0.14 &  0.34\\
         \midrule
         Supervised & .962 & .722 & .873 & .293 & .296 & .500 & .497 & .563 & .896 & .469 & .914 & .734 & .336 & .888 & .114 & .276 & .318 & .273 & .677 & $.558_{\pm.262}$ \\
         \midrule
         Dense FixMatch (E) & \textbf{.976} & .810 & \textbf{.902} & .414 & .424 & .574 & \textbf{.619} & \textbf{.710} & .906 & .575 & \textbf{.935} & \textbf{.778} & \textbf{.544} & \textbf{.924} & \textbf{.560} & \textbf{.699} & \textcolor{red}{.295} & \textbf{.488} & \textbf{.715} & $.676_{\pm.195}$ \\
         Dense FixMatch (I) & .974 & \textbf{.820} & .900 & \textbf{.467} & \textbf{.432} & \textbf{.578} & .603 & .688 & \textbf{.913} & \textbf{.587} & .931 & .769 & .448 & \textbf{.924} & .495 & .667 & \textbf{.471} & .475 & \textcolor{red}{.661} & $.674_{\pm.184}$ \\
         \bottomrule
    \end{tabular}}\\
    \adjustbox{max width=\textwidth}{
    \begin{tabular}{ccccccccccccccccccccccc}
         \toprule
         \textbf{Method} & Bg. &  Plane &  Bicy. &  Bird &  Boat &  Bottle &  Bus &  Car &  Cat &  Chair &  Cow &  Table &  Dog &  Horse &  Motor. &  Person &  Pott. &  Sheep &  Sofa &  Train &  Tv & \textbf{Mean}\\
          \midrule
         \% pixels & 72.45 &  1.09 &  0.83 &  0.15 &  0.94 &  1.57 &  3.46 &  2.60 &  0.58 &  0.65 &  1.25 &  2.08 &  1.13 &  0.58 &  0.90 &  5.27 &  0.56 &  0.34 &  0.49 &  1.72 &  1.35\\
         \midrule
         Supervised & .883 & .702 & .379 & .227 & \textbf{.574} & .443 & .764 & .639 & .336 & .114 & .429 & .215 & .309 & .332 & .542 & .647 & .199 & .511 & .202 & .610 & .307 & $.446_{\pm.206}$ \\
         \midrule
         Dense FixMatch (E) & .890 &  \textbf{.784} &  \textbf{.523} &  \textbf{.639} &  \textcolor{red}{.500} & .555 &  \textbf{.843} &  .684 &  \textbf{.776} &  .188 &  \textbf{.514} &  \textbf{.439} &  .384 &  .388 &  \textbf{.752} &  \textcolor{red}{.622} &  .356 &  \textbf{.576} &  \textbf{.352} &  .743 &  \textbf{.553} &$.574_{\pm.180}$ \\
         Dense FixMatch (I) & \textbf{.899 }&  .740 &  .459 &  \textcolor{red}{.010} &  \textcolor{red}{.559} & \textbf{ .583} &  .833 &  \textbf{.721} &  .431 &  \textbf{.192} &  .469 &  .347 &  \textbf{.418} &  \textbf{.559} &  .693 &  \textbf{.672} &  \textbf{.364} &  \textcolor{red}{.356} &  .333 &  \textbf{.747} &  .545 &$.520_{\pm.213}$ \\
         \bottomrule
    \end{tabular}}
    \caption{Class-wise IoU on \texttt{val} set of Cityscapes (top) when training on a 93 labeled samples data split and on \texttt{val} set of Pascal VOC 2012 (bottom) when training on a 92 labeled samples data split. We show in \textbf{bold} the best result for each class and in \textcolor{red}{red} the classes that perform worse than the supervised baseline for both the \textit{explicit} (E) and \textit{implicit} (I) mini-batch sampling settings.}
    \label{tab:classwise}
    \vspace{-10pt}
\end{table*}

\begin{figure*}[h!]
    \hspace{-18pt}
    \begin{tabular}{ccccc}
        \tiny{Input} & \tiny{Supervised} & \tiny{Dense FixMatch(E)} & \tiny{Dense FixMatch(I)} & \tiny{Ground truth}\\
         \includegraphics[width=.19\textwidth]{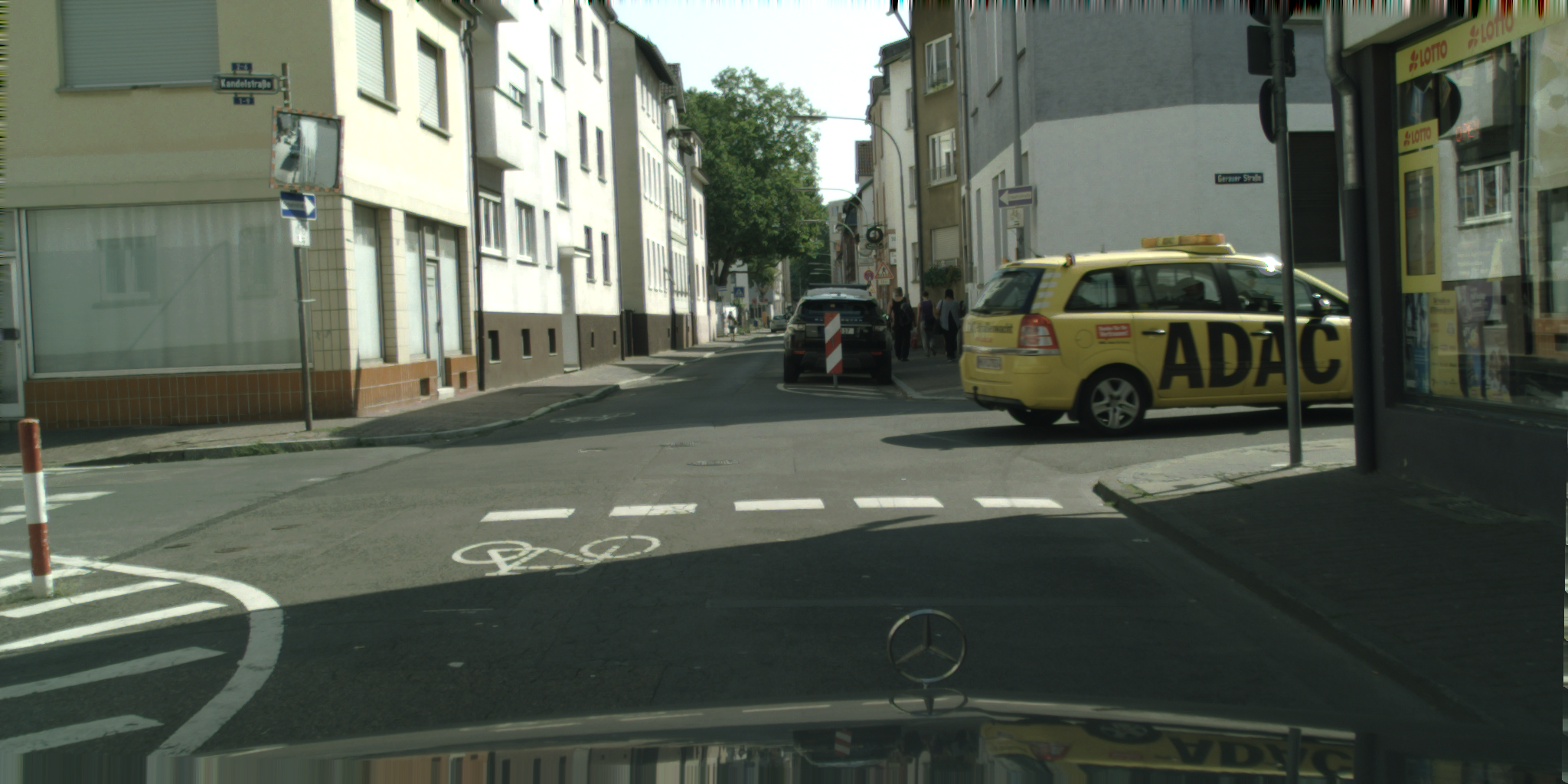} & \includegraphics[width=.19\textwidth]{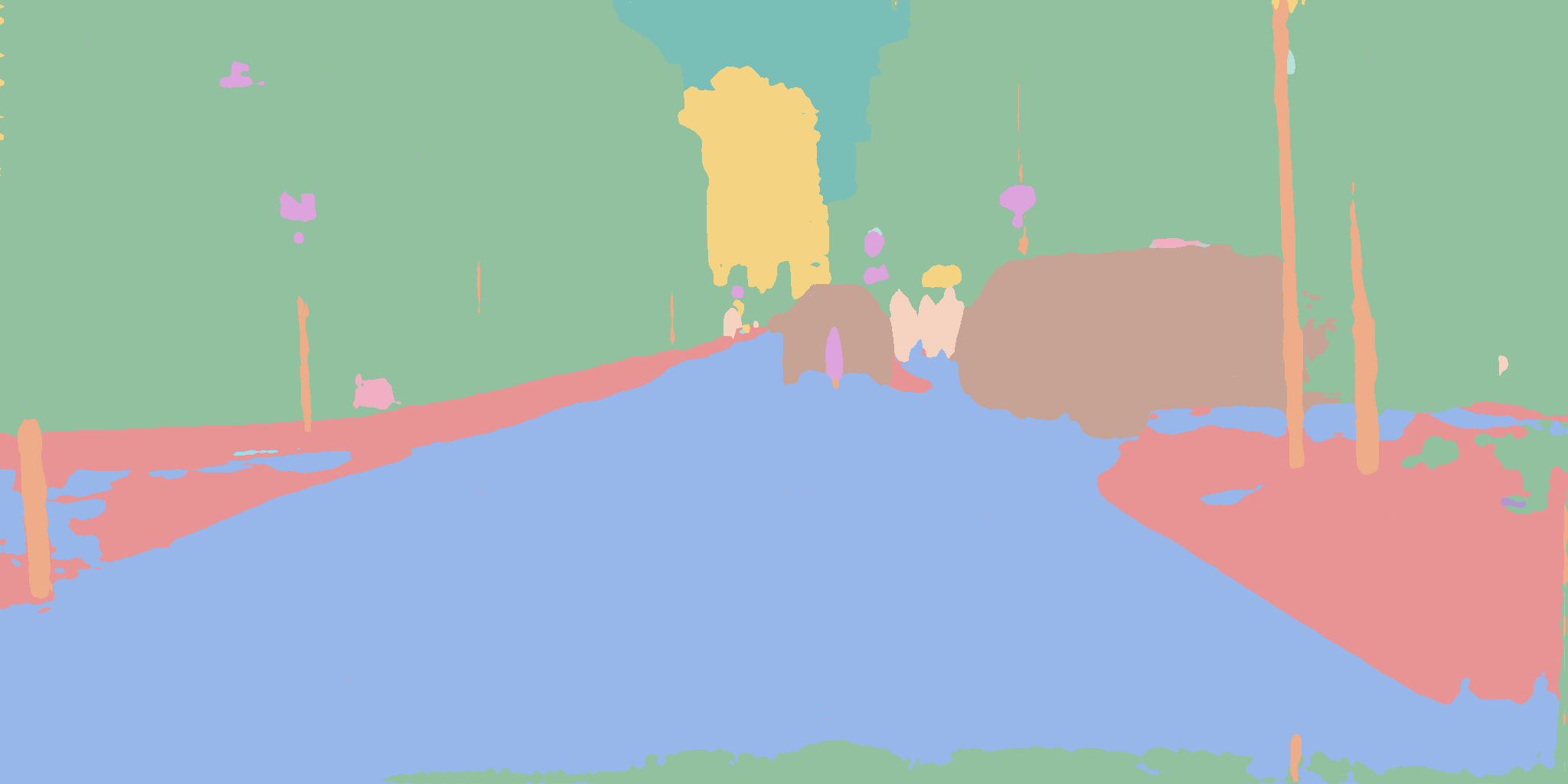} & \includegraphics[width=.19\textwidth]{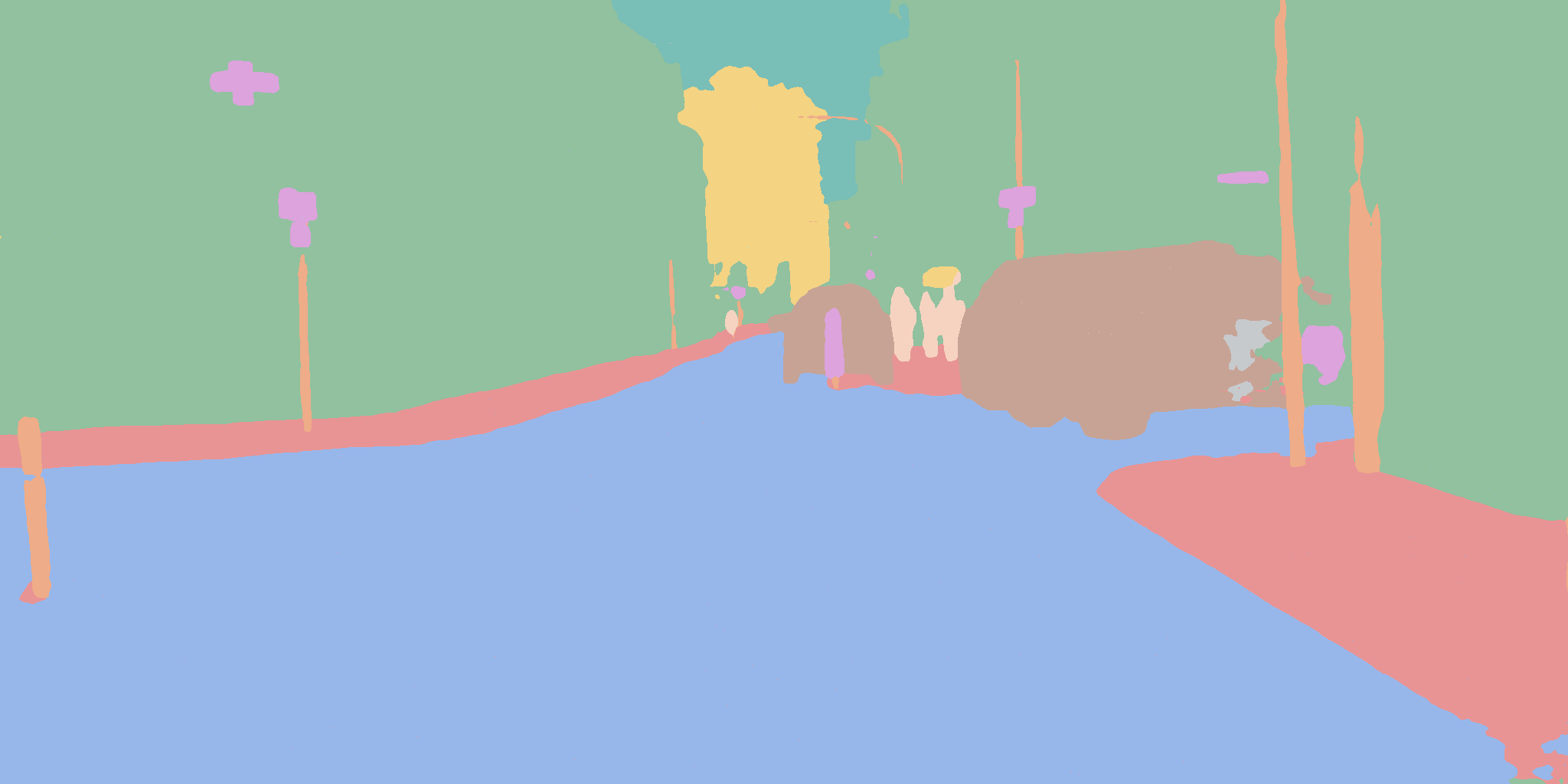} & \includegraphics[width=.19\textwidth]{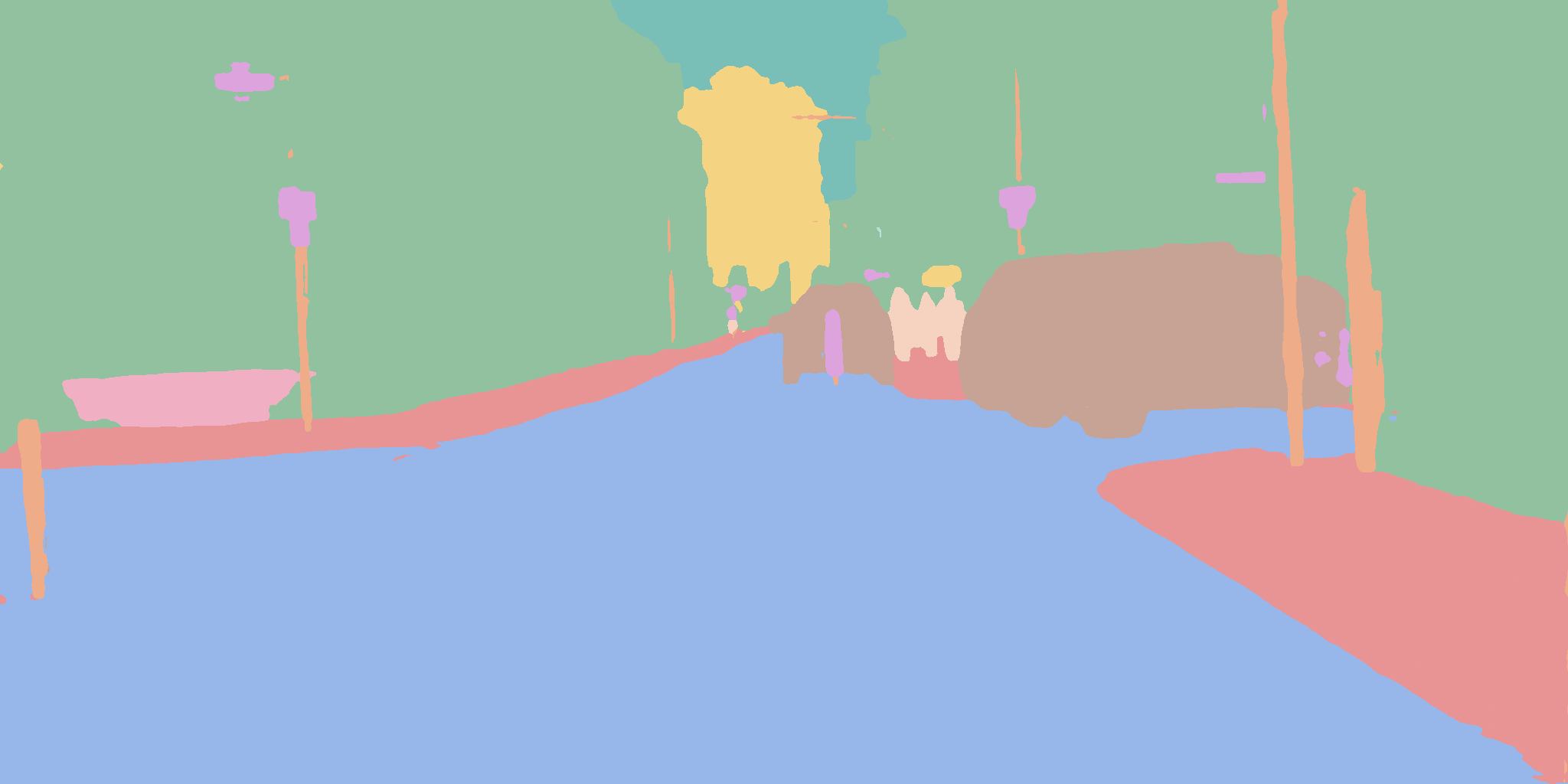} & \includegraphics[width=.19\textwidth]{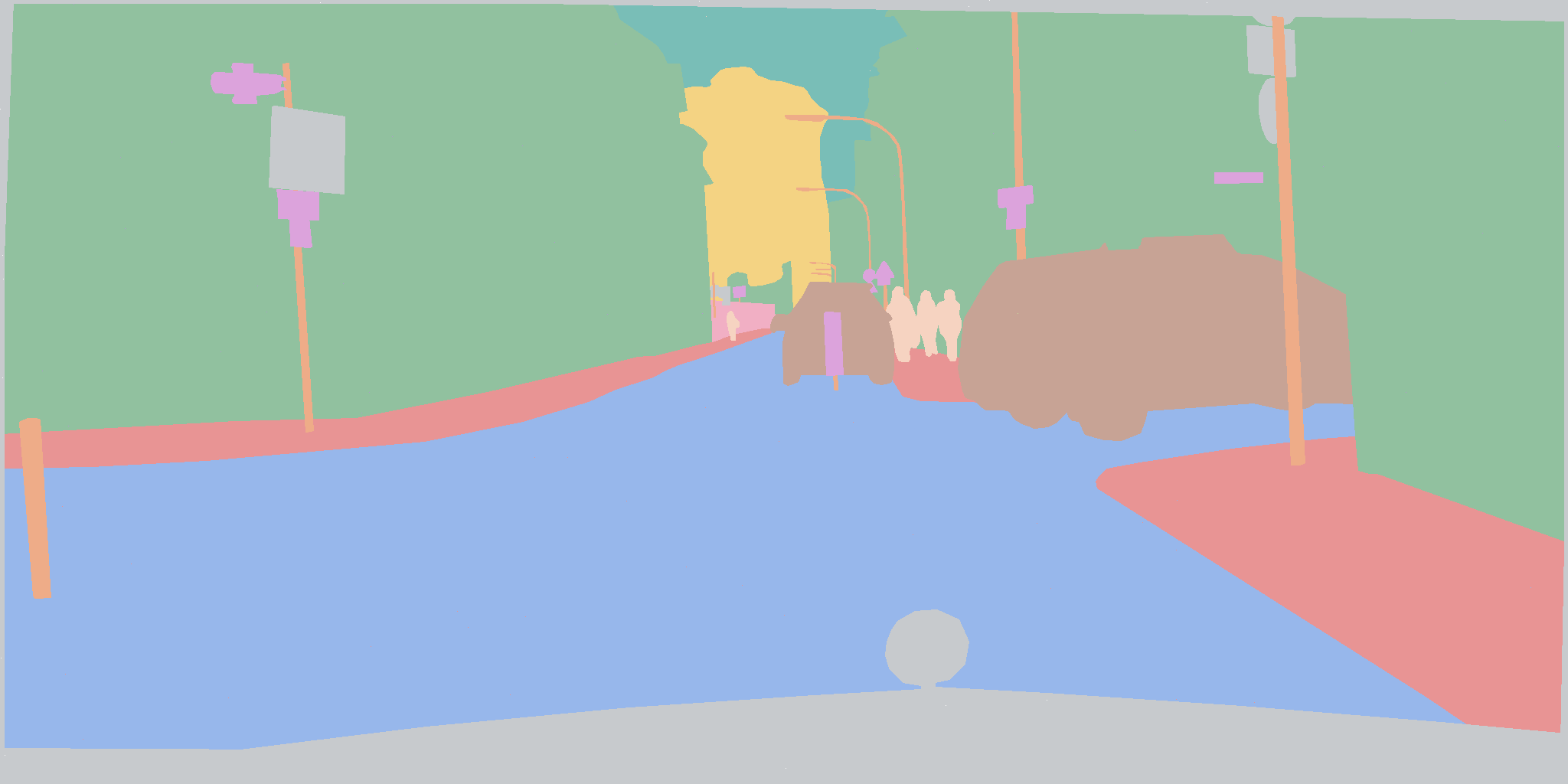} \\
         \includegraphics[width=.19\textwidth]{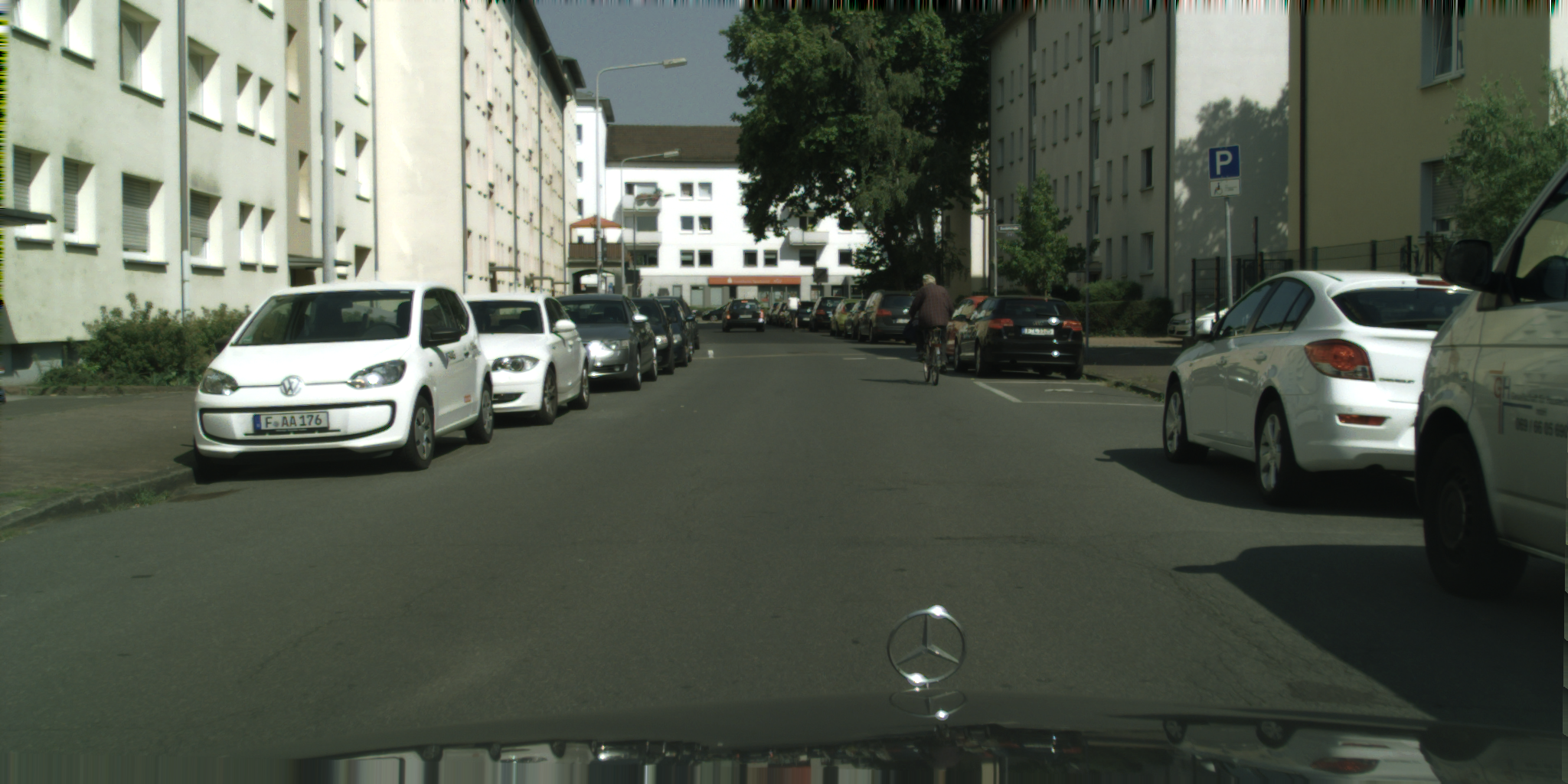} &
         \includegraphics[width=.19\textwidth]{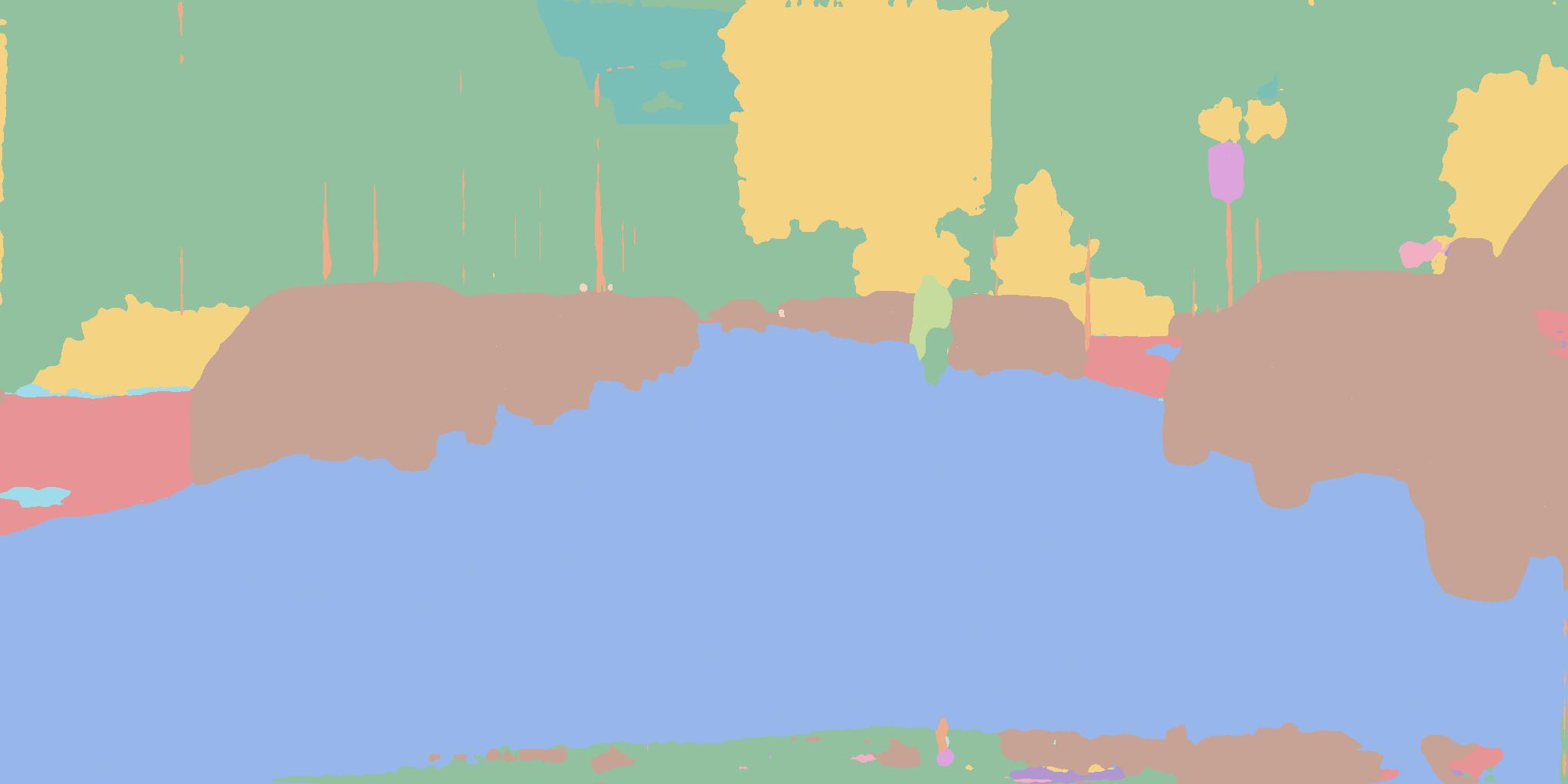} & \includegraphics[width=.19\textwidth]{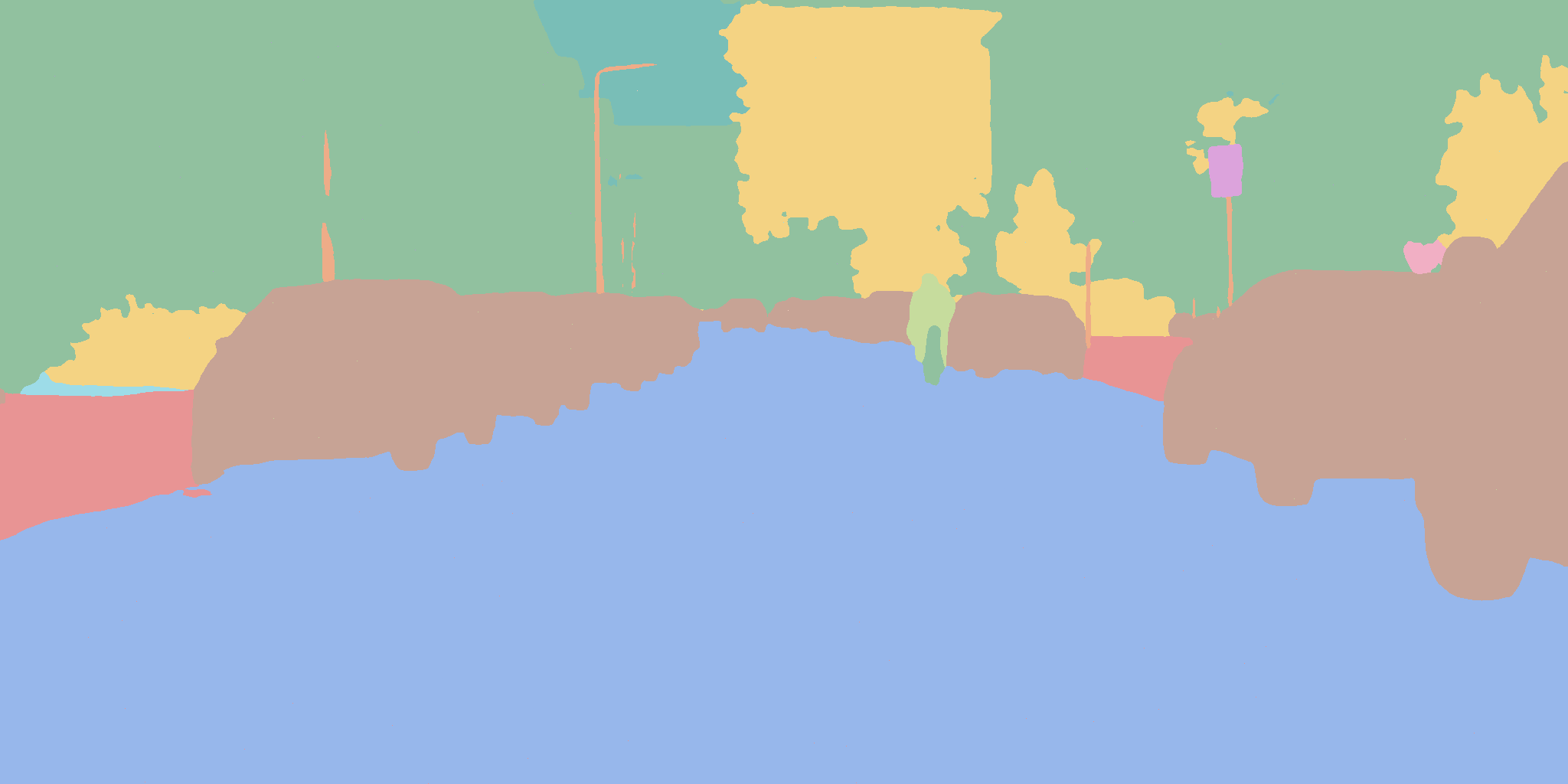} &  \includegraphics[width=.19\textwidth]{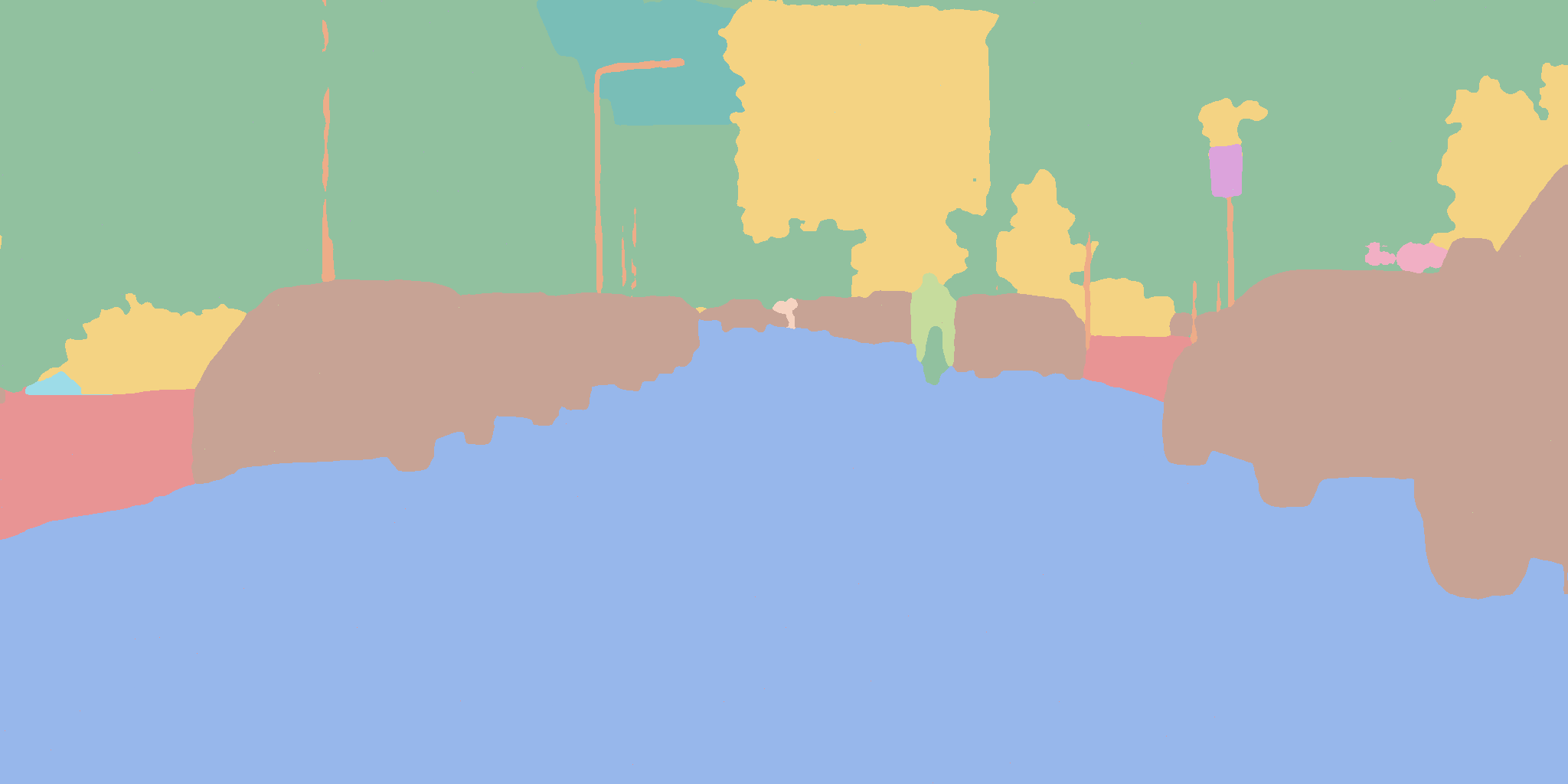} & \includegraphics[width=.19\textwidth]{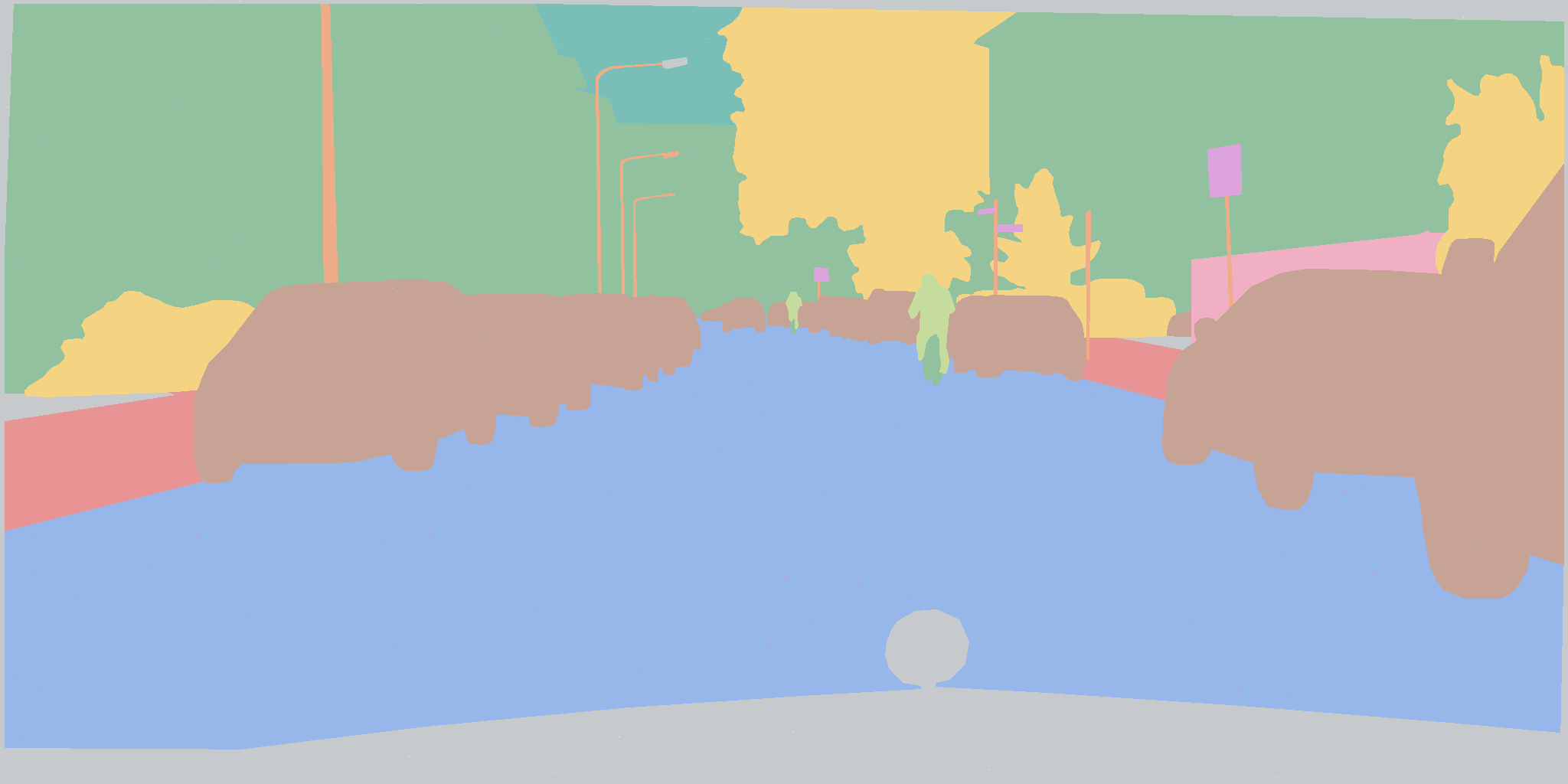} \\
    \end{tabular}
    \caption{Qualitative results for SSL training on Cityscapes with 93 labeled samples for the supervised baseline and Dense FixMatch in the \textit{explicit} (E) and \textit{implicit} (I) mini-batch sampling settings, shown on samples in the validation set.}
    \label{fig:qualitative}
    \vspace{-10pt}
\end{figure*}

\subsection{Class-wise analysis}

Semantic segmentation is known for its large class imbalance. In both Cityscapes and Pascal VOC the predominant classes appear in a few orders of magnitude more pixels than the least frequent ones~\cite{cityscapes,voc}. In Table~\ref{tab:classwise}, we give per-class results comparing the supervised baseline to Dense FixMatch on a single data split of 93 or 92 labeled samples for Cityscapes and Pascal VOC 2012 respectively. We also report the percentage of pixels for each class. For Cityscapes, our method improves significantly on average over the baseline, and does so while improving for all but one class. Importantly, it also reduces the effect of class-imbalance since the gap between the best performing classes and the worst performing ones is reduced significantly, as it can be seen by the lower standard deviation across classes. In Pascal VOC there is a very large imbalance between the background class and the rest. Dense FixMatch improves the results on average, but up to 3 classes get lower IoU. For the implicit setting, the least-frequent class \texttt{Bird} is never learnt.

\subsection{Qualitative results}

In Figure~\ref{fig:qualitative}, we give some examples for the 93 labeled samples experiments with Cityscapes comparing the baseline and Dense FixMatch for both the explicit and implicit settings. We see how both settings give similar results and outperform the supervised baseline, as seen in the more defined boundaries between road and side-walk and poles.

\vspace{-5pt}

\section{Conclusions}
We proposed Dense FixMatch, a simple method that puts together the most important components in modern deep semi-supervised learning and adds a matching operation on the pseudo-labels. In this way, it can be used for multiple dense or structured prediction tasks with the full strength of data augmentation pipelines, including strong geometric transformations.
We evaluated it on semi-supervised semantic segmentation on Cityscapes and Pascal VOC and ablate design choices as well as hyper-parameters. This gives future practitioners insights on how to tune the proposed method for other datasets and tasks.


\section*{Acknowledgments}

This work was partially supported by the Wallenberg AI, Autonomous Systems and Software Program (WASP) funded by the Knut and Alice Wallenberg Foundation.

The computations were partially enabled by resources provided by the Swedish National Infrastructure for Computing (SNIC) at Chalmers Centre for Computational Science and Engineering (C3SE) partially funded by the Swedish Research Council through grant agreement no. 2018-05973.
\bibliographystyle{abbrvnat}
\bibliography{egbib}

\end{document}